\documentclass[10pt,twocolumn,letterpaper]{article}

\usepackage{iccv}
\usepackage{times}
\usepackage{epsfig}

\usepackage{graphicx}
\usepackage{amsmath}
\usepackage{amssymb}
\usepackage{booktabs}
\usepackage[table,xcdraw]{xcolor}

\usepackage{mathtools}

\usepackage[pagebackref,breaklinks,colorlinks]{hyperref}

\usepackage[capitalize]{cleveref}
\crefname{section}{Sec.}{Secs.}
\Crefname{section}{Section}{Sections}
\Crefname{table}{Table}{Tables}
\crefname{table}{Tab.}{Tabs.}

\newcommand{\MethodName}{{Global Knowledge Calibration}}

\iccvfinalcopy % *** Uncomment this line for the final submission

\def\iccvPaperID{5229} % *** Enter the ICCV Paper ID here

\newcommand{\myparagraph}[1]{\smallskip \noindent \textbf{#1}}

\begin{document}

\title{Global Knowledge Calibration for Fast Open-Vocabulary Segmentation}

% \and

\author{Kunyang Han\textsuperscript{1}\footnotemark[1]~,
        Yong Liu\textsuperscript{2}\footnotemark[1]~,
        Jun Hao Liew\textsuperscript{3}~,
        Henghui Ding\textsuperscript{4}~,
        Yunchao Wei\textsuperscript{1}\footnotemark[2]~,\\
        Jiajun Liu\textsuperscript{3}~,
        Yitong Wang\textsuperscript{3}~,
        Yansong Tang\textsuperscript{2}~,
        Yujiu Yang\textsuperscript{2}\footnotemark[2]~,
        Jiashi Feng\textsuperscript{3}~,
        Yao Zhao\textsuperscript{1}~\\
\textsuperscript{1}Beijing Jiaotong University,
\textsuperscript{2}Tsinghua Shenzhen International Graduate School, Tsinghua University,\\
\textsuperscript{3}ByteDance Inc.,
\textsuperscript{4}Nanyang Technological University
}

\maketitle
\footnotetext[1]{Equal contribution.}
\footnotetext[2]{Corresponding author.}

%%%%%%%%% ABSTRACT
\begin{abstract}

Recent advancements in pre-trained vision-language models, such as CLIP, have enabled the segmentation of arbitrary concepts solely from textual inputs, a process commonly referred to as open-vocabulary semantic segmentation (OVS). However, existing OVS techniques confront a fundamental challenge: the trained classifier tends to overfit on the base classes observed during training, resulting in suboptimal generalization performance to unseen classes. To mitigate this issue, recent studies have proposed the use of an additional frozen pre-trained CLIP for classification. Nonetheless, this approach incurs heavy computational overheads as the CLIP vision encoder must be repeatedly forward-passed for each mask, rendering it impractical for real-world applications.
To address this challenge, our objective is to develop a fast OVS model that can perform comparably or better without the extra computational burden of the CLIP image encoder during inference. To this end, we propose a core idea of preserving the generalizable representation when fine-tuning on known classes. Specifically, we introduce a text diversification strategy that generates a set of synonyms for each training category, which prevents the learned representation from collapsing onto specific known category names. Additionally, we employ a text-guided knowledge distillation method to preserve the generalizable knowledge of CLIP. Extensive experiments demonstrate that our proposed model achieves robust generalization performance across various datasets. Furthermore, we perform a preliminary exploration of open-vocabulary video segmentation and present a benchmark that can facilitate future open-vocabulary research in the video domain. 
% Code will be made available. 

\end{abstract}

%%%%%%%%% BODY TEXT
\section{Introduction}
Semantic segmentation aims to group pixels that belong to the same categories. Despite achieving high performance in recent years~\cite{fcn,aspp,unet,segformer, deeplab, maskformer, segnext}, existing semantic segmentation approaches often rely on predefined sets of training categories and thus cannot recognize categories that were not present during training. This limitation greatly restricts their practical applicability.
In contrast, humans possess the ability to recognize novel categories in an open-vocabulary manner, \ie, identifying objects using arbitrary text from an unbounded vocabulary. This ability has inspired the development of open-vocabulary segmentation methods~\cite{zegformer,Simbaseline,pmosr,openseg,spnet,PAD,lseg}. Unlike traditional closed-set segmentation, open-vocabulary segmentation can segment arbitrary categories given only text inputs, which has many potential applications, such as image editing and human-robot interaction.

\begin{figure}[t]
    \centering
    \includegraphics[width=0.95\linewidth]{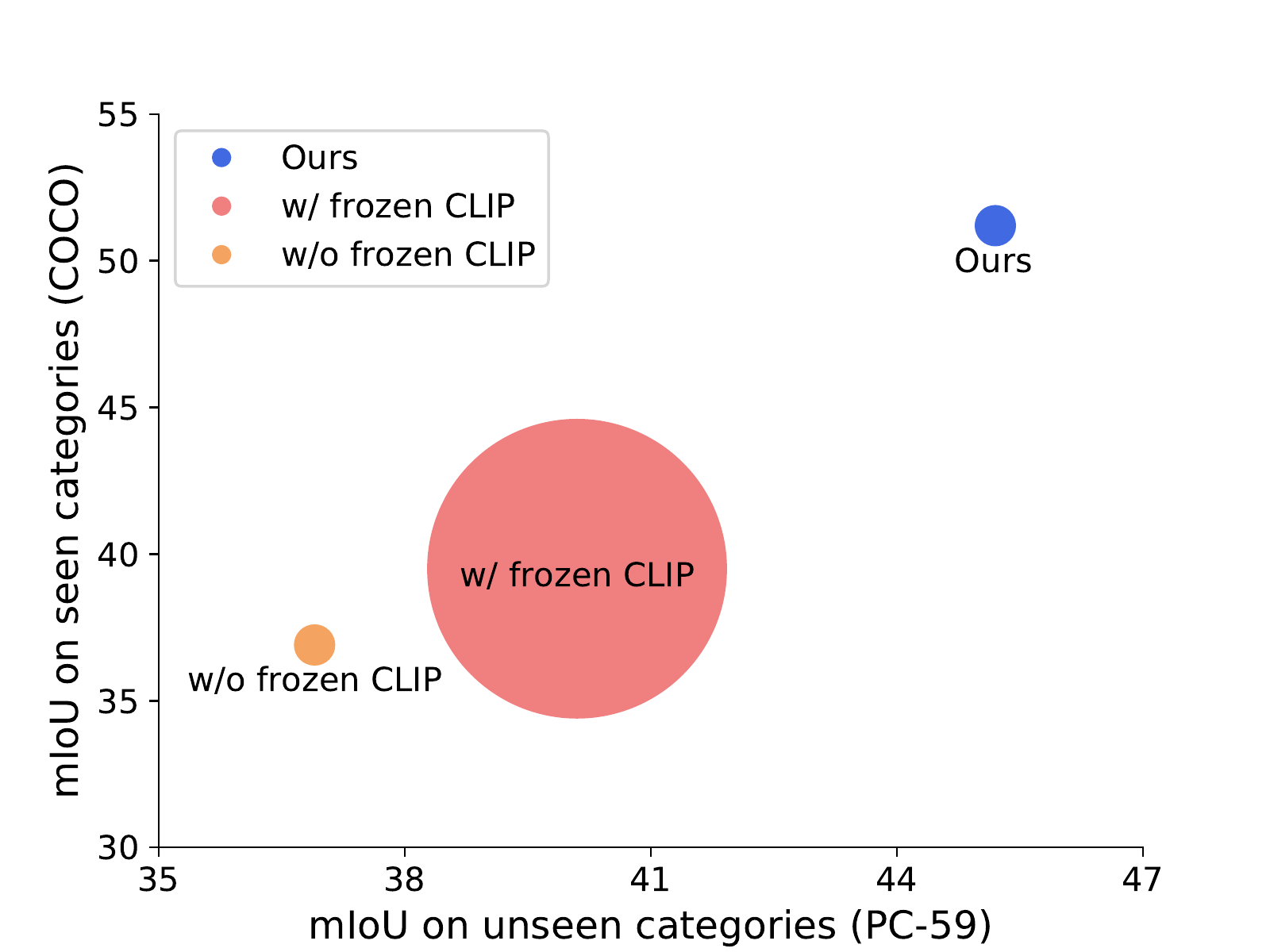}
    \caption{\textbf{Performance \vs computational cost}. The radius of the circle represents the FLOPs during inference. To avoid overfitting to the seen categories, some methods~\cite{zegformer,Simbaseline} introduce an \textbf{extra} frozen CLIP during inference. However, such a strategy leads to heavy computation overhead (\textcolor{red}{red $\bullet$}).
    In comparison, our method generalizes well on both seen and unseen categories with much smaller computational cost (\textcolor{blue}{blue $\bullet$}).
    }
    \vspace{-15pt}
    \label{fig:my_label}
\end{figure}

% \begin{figure}[t]
%     \centering
%     \includegraphics[width=\linewidth]{figures/hist_pc59_flops.pdf}
%     \caption{
%     \textbf{Generalization \vs computational cost}. IoU on unseen categories and FLOPs during inference are reported. To avoid overfitting to the seen categories, some methods~\cite{zegformer,Simbaseline} introduce an \textbf{extra} frozen CLIP. However, such a strategy leads to heavy computation overhead.
%     In comparison, our method generalizes well on unseen categories with much smaller computational costs.
%     }
%     \vspace{-15pt}
%     \label{fig:hist_iou_flops}
% \end{figure}

To achieve open-vocabulary segmentation, early approaches~\cite{spnet,pmosr,lseg} replace the output classification layer with cross-modal alignment, where the similarity measure between pixels and text embeddings is used. Recent works~\cite{zegformer,openseg,adapt-mask,D2Zero,Simbaseline}, on the other hand, adopt the region-level alignment approach and have demonstrated remarkable performance. Despite these advancements, open-vocabulary segmentation methods still face a significant challenge: the learned embeddings often overfit to the base classes observed during training, which hinders their ability to generalize to novel classes. To overcome this challenge, some methods~\cite{Simbaseline,zegformer,adapt-mask} utilize an additional frozen CLIP vision encoder for re-classification. However, this strategy incurs heavy computation overhead, as it requires repeated forward passes of the CLIP vision encoder for each mask. This can be prohibitively expensive for real-world applications, as illustrated in \cref{fig:my_label}.

Therefore, our objective is to train an open-vocabulary semantic segmentation model that is fast and does not require the extra heavy CLIP image encoder during inference, while achieving comparable or better performance. The two main factors that contribute to this objective are: (1) the model should not overfit to the specific training category names, and (2) the model should maintain a feature space similar to the pre-trained CLIP. To achieve this goal, we introduce \textbf{Global Knowledge Calibration}.
To prevent the learned representation from being biased towards the specific training category names, we propose a text diversification strategy for prompt augmentation. This strategy enhances text diversity and enriches category semantics with information of different granularities. Specifically, we use WordNet~\cite{wordnet} to generate a set of synonyms for each training category, \eg, ``vessel'' and ``ship'' for ``boat'', and expand the initial text prompts with this set of words.

To maintain the generalizable knowledge of CLIP~\cite{clip}, a straightforward solution is to apply knowledge distillation. However, traditional knowledge distillation methods only utilize the CLIP features of the same object as supervision. As a result, they can only fit the representations of individual classes and fail to effectively model the overall CLIP space. To address this issue, we propose a text-guided knowledge distillation strategy for calibrating the representation of the trained model. Specifically, we apply distillation supervision for the visual embeddings of one category by using all categories present in the image. Using the distance between category names in the text space as guidance, this distillation strategy can guide the trained model to build a multi-modal feature space similar to the pre-trained CLIP.

In addition, to our best knowledge, previous research on OVS has only focused on the image domain. In this work, we make a preliminary exploration of open-vocabulary video segmentation. We introduce a benchmark by partitioning the large-scale video segmentation dataset, VIPSeg~\cite{vipseg}, into seen and unseen categories for zero-shot testing. We develop a simple baseline based on our image-based method. Our aim is to provide support for future open-vocabulary research in the video domain.

Our contributions can be summarized as follows:
\begin{itemize}
    \item We propose \MethodName\ to preserve generalizable representations when training solely on known classes. Our approach does not require an additional heavy CLIP vision encoder during inference, making it faster. Extensive experiments demonstrate that our model offers strong generalization performance across various datasets, with a much smaller computational cost.
    \item We present a text diversification strategy to enrich text supervision with category information of varying granularities. We propose a text-guided knowledge distillation strategy to calibrate the learned feature space.
    \item To the best of our knowledge, we are the first to explore open-vocabulary video segmentation. We construct a new benchmark and a simple baseline.
\end{itemize}

\begin{figure*}[t]
    \centering
    \includegraphics[width=\linewidth]{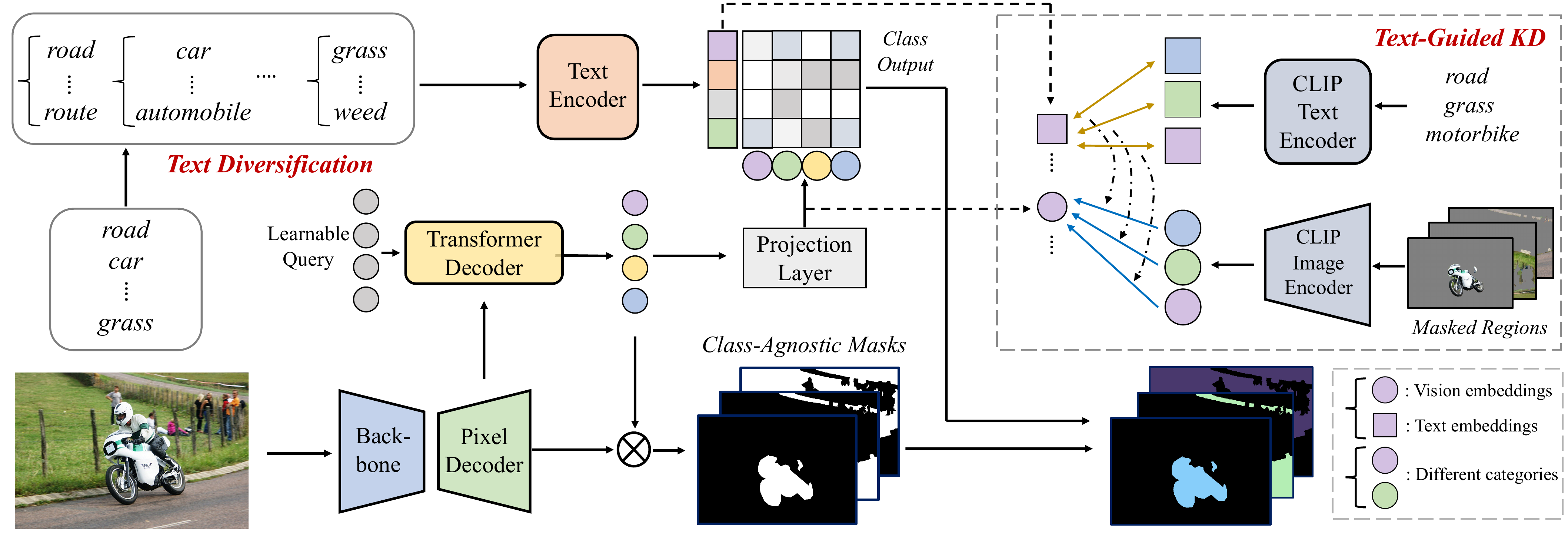}
    % \vspace{-5pt}
    \caption{\textbf{Pipeline}. The input image is first encoded into hierarchical features by visual backbone and pixel decoder. A transformer decoder takes the hierarchical features and a group of learnable queries as input and outputs visual region queries. By perceiving image content, the region queries contain the information of different category regions in the image. By combining the region queries with hierarchical visual features, the model can generate class-agnostic mask proposals. Simultaneously, the region queries are projected towards the textual space with a projection layer. By calculating the similarity between the region queries with the text embeddings of each category name, the model outputs the classification prediction for each mask. During training, we apply both the \textbf{text diversification strategy} and \textbf{text-guided knowledge distillation} to improve the representation of visual and textual embeddings.}
    \label{fig:method}
\end{figure*}

\section{Related Work}

\paragraph{Vision-Language Pre-training.}
Vision-language pre-training aims to learn a joint visual-textual representation space. Early approaches~\cite{pretrain1, pretrain2, pretrain3, pretrain4} were limited to small-scale datasets and required fine-tuning on downstream tasks. With the availability of large-scale web data, recent works~\cite{pretrain5, clip} have demonstrated the benefits of utilizing such data to learn a more robust multi-modal representation space. CLIP~\cite{clip} leverages the idea of contrastive learning to connect images with their corresponding captions and has achieved impressive cross-modal alignment performance. Inspired by previous works~\cite{zegformer, lseg, Simbaseline}, we utilize the well-aligned space of CLIP to enhance open-vocabulary segmentation tasks.

\myparagraph{Open-Vocabulary Segmentation.}
The open-vocabulary segmentation task aims to segment an image and identify regions with arbitrary text queries~\cite{openseg,zs3net}.
Pioneering work by ZS3Net~\cite{zs3net} proposed training a generator to synthesize visual representations by transforming word embeddings.
With the generator expanding the pseudo unseen class visual features, the classifier is trained to distinguish between real features from seen categories and synthetic features from unseen categories.
SPNet~\cite{spnet} replaces the prediction convolution layer by computing the similarity between visual features and linguistic embeddings, while GroupViT~\cite{groupvit} learns to group image regions by contrastive learning between text and images.
LSeg~\cite{lseg} proposes maximizing the correlation between the language embedding and visual pixel-level embeddings using a pre-trained CLIP~\cite{clip} text encoder.
More recently, a two-stage pipeline was proposed: the model first generates class-agnostic region proposals, followed by segment-level alignment between proposals and linguistic embeddings.
OpenSeg~\cite{openseg} leverages a segmentation model to divide input images into regions and computes the grounding loss between the regions and text.
Simbaseline~\cite{Simbaseline} crops the input image based on the proposal masks and utilizes CLIP~\cite{clip} to extract region-level features.
Afterward, the segment embeddings are classified by computing similarity with category name embeddings.
Zegformer~\cite{zegformer} uses CLIP as the encoder and MaskFormer~\cite{maskformer} to extract mask proposals.
However, both Simbaseline and Zegformer require an extra CLIP image encoder to extract the instance embeddings according to the proposal masks, increasing the model parameters and complicating the inference process.
To address these issues and further improve the performance of open-vocabulary segmentation, we propose \MethodName\ in this paper.

\section{Global Knowledge Calibration}
% \subsection{Pipeline}
\myparagraph{Pipeline.}
As depicted in \cref{fig:method}, our method utilizes a ``segment-then-classify'' pipeline for open-vocabulary segmentation task. Initially, the input image is encoded into hierarchical visual features by a visual backbone and a pixel decoder. Subsequently, a transformer~\cite{transformer,detr} decoder takes a set of learnable queries and hierarchical visual features as input to generate region-aware queries (indicated by colored circles in the figure). Next, the region-aware queries are fused with the output of the pixel decoder to produce class-agnostic masks. Concurrently, the region-aware visual queries are fed into a projection layer to perform cross-modal alignment with textual embeddings. The alignment score represents the classification confidence of each query. By combining the class-agnostic masks with the cross-modal alignment scores, our model assigns categories to each mask based on the maximum score. For cross-modal alignment, we use textual embeddings generated by a frozen text encoder~\cite{clip} that takes category names with prompt templates as input. Notably, unlike conventional approaches that rely on the initial class name defined in training datasets, we propose a text diversification strategy to enhance text diversity (\cref{sec:syn}). Specifically, we leverage WordNet~\cite{wordnet} to generate a set of synonyms for each category name, and perform cross-modal matching on all synonyms with corresponding scores. Furthermore, given the high generalizability of pre-trained CLIP~\cite{clip} space, we propose a text-guided knowledge distillation strategy to maintain the CLIP representation even for unseen categories (\cref{sec:kd}).

\subsection{Text Diversification Strategy}\label{sec:syn}

Using only category names as text prompts during training can result in overfitting to specific words and limit the model's ability to generalize. To overcome this limitation, we propose a text diversification strategy that enriches the text prompts with different words that have similar meanings. 
To achieve this, we leverage WordNet~\cite{wordnet} to generate a set of synonyms ${w_i^0, w_i^1, \dots, w_i^{N_i}}$ for each category name $w_i$ in the training set. We manually filter out noisy synonyms with semantic ambiguity, such as ``rock and roll'' for the terrain ``rock'' category, to obtain a precise synonym set. However, while the generated synonyms can be used to describe the whole category, for a specific instance, there may be a more appropriate word to use. For example, ``child'' and ``man'' are both hyponyms of ``person'', but it is not appropriate to use ``child'' to describe someone in their 40s. To address this issue, we introduce a new synonym score metric that measures the distance between a synonym word and a visual instance.

During training, we randomly switch the ground truth text prompt for an instance $Ins_k$ with $i$-th synonym $w_k^i$ from its category's synonym set, using the synonym score as the probability, which is calculated as follows:
\begin{equation}
    S_i = 
    \frac{\exp(\mathcal{R}(Ins_k) \cdot \mathcal{T}(w_k^i))}
    {\Sigma^{N_k}_{j=1}\exp(\mathcal{R}(Ins_k) \cdot \mathcal{T}(w_k^j))}
\end{equation}
where $\mathcal{R}$ is the CLIP~\cite{clip} vision encoder, which takes images cropped by instance masks as input, and $\mathcal{T}$ is the CLIP text encoder, which takes a synonym word from the category's synonym set as input. $N_k$ is the size of the synonym set, and $\cdot$ represents the cosine similarity calculation.
Our text diversification strategy prevents overfitting to specific words and enriches the text prompts with more varied and meaningful synonyms.

\subsection{Text-Guided Knowledge Distillation}
\label{sec:kd}

\begin{figure}[t]
    \centering
    % \footnotesize
    \includegraphics[width=0.85\linewidth]{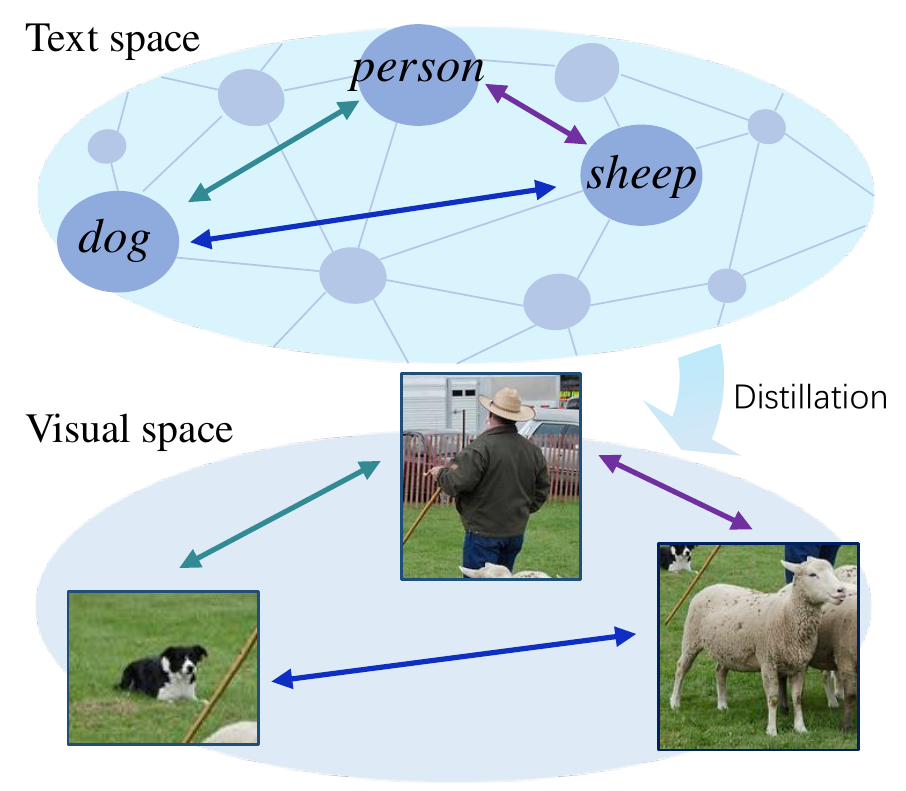}
    % \vspace{-5pt}
    \caption{
    Illustration of text-guided knowledge distillation. Instead of learning a single visual representation from CLIP, our method utilizes the distance among corresponding categories in text space as guidance to learn a structure of various objects in the visual space.
    }
    \label{fig:kd_intro}
\end{figure}

The pre-trained CLIP model is crucial for identifying novel classes and achieving cross-modal alignment. A straightforward approach to leverage CLIP is to incorporate a frozen CLIP image encoder to extract visual embeddings for each mask. Although this approach has shown promising results in recent studies~\cite{Simbaseline, zegformer}, it results in high computation overhead since the CLIP vision encoder must be repeatedly forward passed for each mask proposal, as shown in \cref{tab:cost}. Additionally, since the frozen CLIP encoder is not fine-tuned with known categories, it fails to utilize the training priors to improve recognition of seen categories.

We propose leveraging the well-aligned CLIP space and utilizing knowledge distillation to enhance the generalization ability of visual embeddings. During training, we employ a frozen CLIP image encoder as a teacher model. The teacher model takes images masked by ground truth masks as input and generates region-level visual embeddings for each mask. By imposing constraints between the learned visual queries and the corresponding region embeddings produced by the CLIP teacher, we can take advantage of the superior pre-trained weights of CLIP without increasing the inference process's complexity. This vanilla knowledge distillation can be formulated as:
\begin{equation}
    \mathcal{L}_{KD} = \frac{1}{N}\sum_{i=1}^N\|\mathcal{V}_i - \mathcal{R}(I, M_i)\|,
\end{equation}
where $N$ is the number of ground truth masks in the image. $\mathcal{V}_i$ denotes the generated visual queries matching $i$-th ground truth. $M_i$ is the $i$-th ground truth mask used to mask the image $I$ to serve as input for the CLIP teacher. $\|A - B\|$ denotes the distance measure between $A$ and $B$.

Although the aforementioned vanilla knowledge distillation strategy can fit the representations of individual categories into the CLIP space, it fails to consider the relationships between objects of different categories, making it challenging to build an overall space similar to the pre-trained CLIP.
To overcome this limitation, we propose a \textbf{text-guided} knowledge distillation strategy that utilizes the regions of all categories present in the image to calibrate the representation space of the trained model. As illustrated in \cref{fig:kd_intro}, in a well-aligned CLIP space, the relationship between the visual representations of different categories should be consistent with the relationship between the corresponding texts. Therefore, we can use the distance between category names in the text space as a guidance signal for distilling visual embeddings.

For instance, taking the ``bus'' and ``bear'' as examples, when we distill the student features belonging to the ``bus'' class, the distance between the student visual features and the teacher CLIP features of the ``bear'' region should be the same as the distance between the text embeddings of ``bus'' and ``bear''. The text-guided knowledge distillation process can be formulated as:
\begin{equation}\small
    \mathcal{L}_{TGKD} = \frac{1}{N}\sum_{i=1}^N\sum_{j=1}^N\Big\|(\|\mathcal{V}_i - \mathcal{R}(I, M_j)\| - \|\mathcal{T}(Y_i) - \mathcal{T}(Y_j)\|)\Big\|,
\end{equation}
where $\mathcal{T}$ denotes the CLIP text encoder and $Y_i$ is the category name of the $i$-th ground truth region.

\subsection{Loss Functions}
The total loss consists of three parts: segmentation loss $\mathcal{L}_{M}$, alignment loss $\mathcal{L}_{A}$, and knowledge distillation loss $\mathcal{L}_{TGKD}$. To supervise the output mask proposals, we adopt a combination of binary cross-entropy loss and dice loss~\cite{dice-loss}, following query-based segmentation methods~\cite{maskformer,cheng2021mask2former}.
For the alignment loss $\mathcal{L}_{A}$, we utilize cross-entropy to supervise the matching scores. Additionally, we incorporate the grounding loss, following prior work~\cite{openseg, groundingloss1, groundingloss2}, to leverage the image-level captions and encourage region-word alignments. Specifically, the grounding loss maximizes the similarity score of the labeled image-caption pair over all images and all captions in a mini-batch~\cite{openseg}.
The total loss is formulated as follows:
\begin{equation}
    \mathcal{L} = \lambda_m\mathcal{L}_{M} + \underbrace{\lambda_c\mathcal{L}_{CE} + \lambda_g\mathcal{L}_{G}}_{\mathcal{L}_A} + \lambda_{kd}\mathcal{L}_{TGKD},
\end{equation}
where $\lambda$ represents the weight of each loss. $\mathcal{L}_{CE}$ is cross-entropy loss. $\mathcal{L}_{G}$ denotes the grounding loss.

\subsection{Open-Vocabulary Video Segmentation}\label{sec:vipseg_setting}
In order to expand the open-vocabulary task to a broader range of applications, we conduct a preliminary exploration of open-vocabulary video segmentation. Specifically, we divide the large-scale video segmentation dataset VIPSeg~\cite{vipseg} into seen and unseen categories and construct a baseline using our method.

We follow the common video task~\cite{Yang2019vis,mmdetection,heo2022vita} training strategy, which first pre-trains the model on COCO~\cite{coco} and then finetunes on the video dataset. To prevent category information leakage from COCO pre-training, categories of VIPSeg are progressively verified. To this end, we recruited four participants who were asked to recognize the pattern of VIPSeg categories on COCO samples and split VIPSeg categories into three sets. The first set contains categories that are annotated by both datasets and have the same category definition, while the second set contains categories that are annotated by both datasets but differ in the level of granularity, \eg, ``ball net'' and ``goal'' \vs ``net''. The third set includes categories that are either treated as background or not found in COCO samples, such as ``tyre''. Finally, we select 12 categories from the third set as novel categories, which cover a total of 9 super-categories defined by VIPSeg.

Similar to Video Mask2Former~\cite{video_mask2former}, we represent the entire video sequence as a 3D spatio-temporal volume of dimensions $T\times H\times W$, where $T$ is the number of frames, $H$ and $W$ are the height and width, respectively. By extending our approach in a manner similar to Video Mask2Former, our method can be easily adapted to the video scenario.

\section{Experiment}

\begin{table*}[t]
    \centering
    \small
    \renewcommand\arraystretch{1.1}
    \caption{The open-vocabulary segmentation performance comparison on the popular image segmentation datasets. PAS denotes the Pascal VOC~\cite{pascal-voc} dataset and PC denotes the Pascal Context~\cite{pascal} dataset. The best results of each dataset are \textbf{bolded}. The second best results are \underline{underlined}.
    The results on the COCO dataset demonstrate the segmentation ability for the training seen categories. Results on other datasets show the open-vocabulary segmentation ability. $\dag$ denotes the reproduced result using the same training setting for fair comparison.}
    \label{tab:results}
    \setlength\tabcolsep{4pt}
    % \resizebox{\linewidth}{!}{
    \begin{tabular}{l|ll|ccccccc}
        \toprule[1pt]
        Model &Backbone &Training Set &COCO &PAS-20 & Cityscapes& ADE20K-150 & ADE20K-847 & PC-59 & PC-459\\
        \hline
        ZS3Net~\cite{zs3net} &R-101 &PASCAL-15 &- &38.3 &- &- &- &19.4 &-\\
        SPNet~\cite{spnet} &R-101 &PASCAL-15 &- &18.3 &- &- &1.6 &24.3 &-\\
        LSeg~\cite{lseg} &R-101 &PASCAL-15 &- &47.4 &- &- &- &- &- \\
        ZegFormer~\cite{zegformer} &R50 &COCO Stuff &- &\underline{80.7} &- &16.4 &- &- &-\\
        Simbaseline~\cite{Simbaseline} &CLIP R-101  &COCO Stuff &- &74.5 &- &15.3 &- &- &-\\
        LSeg+~\cite{openseg} &R-101 &COCO Panoptic &- &59.0 &- &13.0 &2.5  &36.0 &5.2\\
        OpenSeg~\cite{openseg} &R-101 &COCO Panoptic &36.9 &60.0 &- &15.3  &\textbf{4.0}  &36.9 &6.5\\
        Simbaseline$^{\dag}$~\cite{Simbaseline} &CLIP R-50 &COCO Panoptic &39.5 &- &30.0 &14.4 &- &40.1 &\underline{6.7}\\
        \hline
        Ours &CLIP R-50 &COCO Panoptic &\underline{49.8} &78.7 &\underline{34.3} &\underline{17.5}  &3.2  &\underline{41.9} &6.5\\
        Ours &R-101 &COCO Panoptic &\textbf{51.2} &\textbf{83.2} &\textbf{34.8} &\textbf{18.8}  &\underline{3.5}  &\textbf{45.2} &\textbf{7.1}\\
        % Ours &R-50c &COCO Panoptic &- &- &-  &-  &- &-\\
        
        \bottomrule[1pt]
    % \vspace{-15pt}
    \end{tabular}
    % }
    \end{table*}

Following previous works~\cite{Simbaseline,openseg,zegformer}, we evaluate our open-vocabulary image segmentation model in a cross-dataset setting. In this setting, the model is trained on one dataset and evaluated on other datasets without fine-tuning or retraining. The default training dataset in this paper is COCO Panoptic~\cite{coco} with 133 categories, as used in previous works~\cite{openseg}. This setting is particularly challenging as the model has to handle both unseen classes and domain gaps between different datasets~\cite{Simbaseline}.

\smallskip
\noindent \textbf{Open-Vocabulary Video Segmentation Setting.}
Due to the limited number of large-scale video segmentation datasets, we evaluate our open-vocabulary video segmentation model in the ordinary zero-shot setting. In this setting, the model is trained on the seen categories and evaluated on both seen and unseen categories.

\subsection{Datasets and Evaluation Metrics}
To evaluate the effectiveness of our method, we conduct extensive experiments on the image and video datasets, COCO~\cite{coco}, ADE20K~\cite{ade20k}, Cityscapes~\cite{cityscape},  Pascal VOC 2012~\cite{pascal-voc}, Pascal Context~\cite{pascal}, and VIPSeg~\cite{vipseg}.\\
\textbf{COCO} is a large-scale dataset with 117k training images and 5k validation images.
We use its panoptic and caption annotations during our training stage, and evaluate it in semantic segmentation manner.\\
\textbf{ADE20K} contains 20k training images, 2k validation images, and 3k testing images. 
There are two splits of this dataset. 
ADE20K-150 contains 150 semantic classes whereas ADE20K-857 has 857 classes.
In this paper, we take both splits to verify the performance of our method.\\
\textbf{Cityscapes} is a scene parsing dataset with 5,000 accurately annotated images and 20,000 coarsely annotated images. 
Following previous works~\cite{cityscape, Simbaseline}, we take 1,525 images of 19 classes in the accurately annotated set for validation.\\
\textbf{Pascal VOC 2012} contains 11,185 training images and 1,449 validation images from 20 classes. We use the provided augmented annotations.\\
\textbf{Pascal Context} is an extension of Pascal VOC 2010, containing 4,998 training images and 5,005 validation images. 
In this paper, we take the commonly used PC-59 and challenging PC-459 version for validation.\\
\textbf{VIPSeg} contains 124 classes, including 3,536 videos and 84,750 frames with pixel-level panoptic annotations.\\
\textbf{Evaluation Metric.} Following previous works~\cite{Simbaseline,zegformer,openseg}, we take the \textit{mean-intersection-over-union} (mIoU) as the metric to measure the segmentation performance.
For video datasets, we apply the mIoU on seen classes, unseen classes, and their harmonic mean as major metric.

\subsection{Implementation Details}
Our implementation is based on {\tt detectron2}~\cite{wu2019detectron2}.
All image-based models are trained with batch size of 112 and training iteration of 50k. The base learning rate is 0.0003 with a step schedule, in which the steps are set to 40k and 45k, the scaling factor is 0.1. The input image is resized to 512$\times$512. For data augmentation, random horizontal flip and multi-scale jittering with a random scale between [0.8, 1.2] are applied.
For the weights of the loss function, we set $\lambda_m$ to 5, $\lambda_c$, $\lambda_g$ and $\lambda_{kd}$ to 2 by default.
All video-based models are fine-tuned with batch size of 16 and training iteration number of 3k. The base learning rate is 0.0001 with a step schedule, and the step is set to 2k.
The backbones of both our model and the distillation teacher CLIP model are CLIP ResNet-50~\cite{clip} by default. Note that the text encoder of CLIP is frozen in the training stage. Other hyperparameters are the same as Mask2Former~\cite{cheng2021mask2former}.

\subsection{Comparison with State-of-the-Art Methods}
We evaluate the effectiveness of our proposed method against state-of-the-art techniques on several popular image segmentation datasets~\cite{coco, ade20k, pascal, pascal-voc, cityscape}, to assess its open-vocabulary performance. The results are presented in \cref{tab:results}. The obtained results indicate that our method demonstrates strong open-vocabulary segmentation ability. Specifically, when trained on the COCO Panoptic~\cite{coco} dataset, which contains 133 categories, our method achieves 7.1 mIoU on the complete Pascal Context~\cite{pascal} dataset with 459 categories and 18.8 mIoU on the ADE20K~\cite{ade20k} dataset with 150 categories. Moreover, our approach, which utilizes CLIP ResNet-50 as the backbone, outperforms prior work utilizing CLIP ResNet-101. The comparison on COCO~\cite{coco} dataset verifies the effectiveness of our method for in-domain segmentation tasks. Compared with previous approaches, our method shows remarkable open-vocabulary segmentation capability while significantly improving the recognition of training categories. Additionally, we perform experiments with ImageNet~\cite{imagenet} pre-trained backbone, and our model also achieves promising results, demonstrating the flexibility of our approach.

\begin{table}[t]
    \centering
    \small
    \renewcommand\arraystretch{1.1}
    \caption{Computational cost comparison between our method and current two-stage methods. The FLOPs and Params are measured on the backbone of ResNet101. The FPS is recorded on the same single V100 GPU.}
    \label{tab:cost}
    \setlength\tabcolsep{11.2pt}
    % \resizebox{\linewidth}{!}{
    \begin{tabular}{l|ccc}
        \toprule[1pt]
        Model &FLOPs &Params &FPS\\
        \hline
        Simbaseline~\cite{Simbaseline} &1165.07G  &89.76M &2.32\\
        ZegFormer~\cite{zegformer} &1127.86G  &63.90M &5.39\\
        Ours &151.44G  &40.51M &8.04\\
        % Ours &256.82G  &64.03M &16.67\\
        \bottomrule[1pt]
    \end{tabular}
    % }
\end{table}

We also provide a comparison of the computational complexity and efficiency of our method with two previous two-stage methods~\cite{zegformer, Simbaseline}. As shown in Table~\ref{tab:cost}, existing region-level alignment methods require an additional frozen CLIP~\cite{clip} vision encoder to extract foreground visual features for each mask proposal, leading to massive models and slower inference speeds. In contrast, our method achieves high segmentation performance while maintaining a reasonable computation cost. The table shows that our method has approximately 10\% of the FLOPs of the previous methods, and a significant increase in FPS can also be observed.

\begin{table}[t]
    \centering
    \small
    \renewcommand\arraystretch{1.1}
    \caption{The quantitative results of the video open-vocabulary segmentation. The model is trained on the seen categories and evaluated on both the seen and unseen categories.}
    \label{tab:video_results}
    \setlength\tabcolsep{12.5pt}
    % \resizebox{\linewidth}{!}{
    \begin{tabular}{lccc}
        \toprule[1pt]
        Model  &Seen &Unseen &Harmonic\\
        \hline
        Baseline &44.2 &2.4 &4.5\\
        + KD &43.4 &2.9 &5.4\\
        + KD + TD &45.8 &8.5 &14.4\\
        
        \bottomrule[1pt]
    % \vspace{-15pt}
    \end{tabular}
    % }
\end{table}

\subsection{Open-Vocabulary Video Segmentation}
Recognizing novel categories in videos is a challenging task due to the complexity and variability of video scenes. The segmentation results of our proposed method on the VIPSeg~\cite{vipseg} dataset are presented in \ref{tab:video_results}. Following Video Mask2Former~\cite{video_mask2former}, we extend our method to a video version and finetune it on the seen categories. The baseline in \ref{tab:video_results} refers to the model trained with mask loss, cross-entropy loss, and dice loss~\cite{dice-loss} only. The corresponding mIoU values of the seen, unseen, and harmonic categories are 44.2, 2.4, and 4.5, respectively.
With the addition of text-guided knowledge distillation supervision, the mIoU values of the unseen and harmonic categories improved by 0.5 and 0.9, respectively. Moreover, by utilizing our proposed text diversification strategy, the model is able to achieve 8.5 and 14.4 on unseen and harmonic mIoU, respectively, which is almost three times improvement over the baseline method.

\subsection{Ablation Study}
In this section, we conduct several ablations to justify the design choices in our proposed network.

\myparagraph{Component Analysis.}
To verify the effectiveness of our proposed strategies, we conduct experiments on the Pascal Context~\cite{pascal} and Cityscapes~\cite{cityscape} datasets.
The results are shown in \cref{tab:component}.
In the table, TD denotes the text diversification training strategy, and TGKD denotes the proposed text-guided knowledge distillation.
The model is trained on the COCO Panoptic dataset~\cite{coco} with CLIP ResNet-50 as the backbone.
As can be seen from the results, the text diversification strategy improves the performance by about 2\% and 4\% on Pascal Context and Cityscapes, respectively.
The text-guided knowledge distillation also contributes to a performance gain of 1.6\% and 4\% on these datasets.
When both strategies are used together, the final performance is boosted to 41.91\% and 34.35\%, respectively.

\begin{table}[t]
    \centering
    \small
    \caption{The ablation study on the proposed component. TD denotes the text diversification strategy. TGKD is the text-guided knowledge distillation strategy.}
    \label{tab:component}
    \setlength\tabcolsep{13pt}
    % \resizebox{\linewidth}{!}{
    \begin{tabular}{cc|cc}
        \toprule[1pt]
        TD  &TGKD &Pascal Context &Cityscapes\\
        \hline
         % & &33.33 &13.66 &30.25\\
         & &39.70 &27.61\\
        \checkmark & &41.45 &32.16\\
        & \checkmark  &41.23 &32.62\\
        \checkmark &\checkmark &41.91 &34.35\\
        
        \bottomrule[1pt]
    % \vspace{-15pt}
    \end{tabular}
    % }
\end{table}

\begin{table}[t]
    \centering
    \small
    \caption{Experiment results of different distillation strategies. Here $\times$ indicates that no knowledge distillation is performed. Vanilla denotes the distillation guided by visual features from one ground truth region. Vision-guided means taking visual embeddings from all ground truth regions as supervision.}
    \label{tab:kd_strategy}
    \setlength\tabcolsep{11pt}
    % \resizebox{\linewidth}{!}{
    \begin{tabular}{l|cc}
        \toprule[1pt]
        Distillation Strategy &Pascal Context &Cityscapes\\
        \hline
        $\times$  &39.70 &27.61\\
        Vanilla &40.14 &32.49\\
        Vision-guided  &39.67 &32.33 \\
        Text-guided &41.23 &32.62 \\
        \bottomrule[1pt]
    \end{tabular}
    % }
\end{table}

% \begin{table}[t]
%    \centering
%    \small
%   \renewcommand\arraystretch{1.1}
%   \setlength{\tabcolsep}{4pt}
%    \begin{tabular}{l|c|c|c}
%       \toprule[1pt]
%       & IoU & P@0.5  & P@0.9\\
%       \hline
%       % \multicolumn{4}{l}{ \textit{(a) Leanrnable Query}} \\
%       % \hline
%       % Sentence Embedding & 71.67 & 82.80 & 21.91  \\
%       % + Learnable Query & 70.47 & 81.17 & 20.80 \\
%       % \hline
%       \rowcolor{gray!18}\multicolumn{4}{l}{ \textit{(a) Structure of Language-to-Vision Decoder}} \\
%       \hline
%       1 Decoder Layer  & 71.38 & 82.36 & 21.40 \\
%       3 Decoder Layers  & 71.45 & 82.43 & 21.33   \\
%       6 Decoder Layers  & 71.67 & 82.80 & 21.91   \\
%       + Encoder Layers  & 71.67 & 82.92 & 21.93   \\
%       \hline
%       \rowcolor{gray!18}\multicolumn{4}{l}{ \textit{(b) Structure of Vision Projection Module}} \\
%       \hline
%       Only  Cross-Attention Fusion & 68.55 & 78.64 & 19.38   \\
%       Both Self and Cross-Attention Fusion & 71.67 & 82.92 & 21.93   \\
%       \bottomrule[1pt]
%    \end{tabular}
%    \caption{Some ablation studies about the model architecture on the RefCOCO validation set. The vision encoder used is CLIP-ResNet50~\cite{CLIP}.}
%    \label{tab:some}
%    \vspace{-15pt}
% \end{table}
\myparagraph{Distillation Methods.}
We compared different knowledge distillation methods in \cref{tab:kd_strategy}.
For each generated region query, vanilla distillation constrains it using only the CLIP visual features of its corresponding ground truth region.
This approach does not take into account the relationship between the region query and embeddings of other categories, which may compromise the effect of multi-modal alignment.
To alleviate this problem, we propose to leverage all regions in the image to supervise each visual query.
Since regions have different visual content information, only using the distance between visual embeddings as distillation guide may introduce errors.
Thanks to CLIP's excellent pre-trained common space, the generated queries can learn high-level semantic information for each category by using the distance between text embeddings of different categories as guidance.
Experiment results also prove that using text distance as guidance works best.

\myparagraph{Distillation Features.}
\begin{table}[t]
    \centering
    \small
    \caption{Experiments of different teacher and student embeddings.}
    \label{tab:kd_position}
    \setlength\tabcolsep{8pt}
    % \resizebox{\linewidth}{!}{
    \begin{tabular}{cc|cc}
        \toprule[1pt]
        Teacher  &Student &Pascal Context &Cityscapes\\
        \hline
         % & &33.33 &13.66 &30.25\\
        global token &post &40.71 &30.43\\
        global token &prior &40.31 &31.13\\
        spatial token &post  &41.29 &30.12\\
        spatial token &prior &41.23 &32.62\\
        \bottomrule[1pt]
    % \vspace{-15pt}
    \end{tabular}
    % }
\end{table}
There are various options of teacher embeddings and student embeddings for knowledge distillation.
Specifically, the teacher embedding can be the global token or spatial tokens with mask-based pooling in the attention pooling process of CLIP~\cite{clip}.
For student embedding, we have experimented with the visual queries before and after the projection layer.
The results are shown in \cref{tab:kd_position}.
We find that the best choice is to use spatial tokens with mask-based pooling as the teacher embedding and the queries before projection layer as the student embedding.

\myparagraph{Text Diversification Strategy.}
We experiment with different text diversification strategies in \cref{tab:syn-backbone} (a), including (1) randomly replacing the GT with synonym with probability being its synonym score described in \cref{sec:syn}, (2) taking the maximum (GroupMax) or (3) the average among the complete synonym set as the prediction of the corresponding category. All methods are equally effective, showing that text diversification method is robust to different strategies.
Technically, our proposed text diversification is general and applicable to other open-vocabulary segmentation methods. We additionally verify its effectiveness by applying to Simbaseline~\cite{Simbaseline}, \cref{tab:simbaseline-td} shows that it improves by 2.5\% and 4.7\% mIoU on Pascal Context and Cityscapes, respectively.

\myparagraph{Different Visual Backbones.}
We also experiment with different visual backbones for our method.
As \cref{tab:syn-backbone} shows, the CLIP pre-trained backbones perform better due to the well-aligned multi-modal space.
However, with the proposed text diversification and text-guided knowledge distillation strategies, our method with ImageNet~\cite{imagenet} pre-trained backbones performs equally well. This greatly expands the flexibility of our method since we are no longer constrained to vision-language pre-trained backbones.

\begin{table}[t]
    \centering
    \small
    % \renewcommand\arraystretch{1.1}
    % \caption{Experiment results of different synonym utilization methods. Random denotes xxx. ClusterAvg indicates xxx. ClusterMax denotes xxx.}
    \caption{Experiment results of different text diversification methods and backbones.}
    \label{tab:syn-backbone}
    \setlength\tabcolsep{14pt}
    % \resizebox{\linewidth}{!}{
    \begin{tabular}{l|cc}
        \toprule[1pt]
         &Pascal Context &Cityscapes\\
        \midrule
        \rowcolor{gray!18}\multicolumn{3}{l}{ \textit{(a) Different Text Diversification Method}} \\
        \midrule
        Random &41.45 &32.16\\
        GroupAvg  &41.80 &31.42\\
        GroupMax &41.17 &32.73\\
        \midrule
        \rowcolor{gray!18}\multicolumn{3}{l}{ \textit{(b) Different Visual Backbones}} \\
        \midrule
        ImageNet-R50  &45.6 &32.9\\
        CLIP-R50  &41.9 &34.3\\
        ImageNet-R101  &45.2 &34.8\\
        CLIP-R101  &44.2 &37.6\\
        \bottomrule[1pt]
    \end{tabular}
    % }
\end{table}

\begin{table}[t]
    \centering
    \small
    % \renewcommand\arraystretch{1.1}
    % \caption{Experiment results of different synonym utilization methods. Random denotes xxx. ClusterAvg indicates xxx. ClusterMax denotes xxx.}
    \caption{Experiment results of applying TD to other methods.}
    \label{tab:simbaseline-td}
    \setlength\tabcolsep{14pt}
    \resizebox{\linewidth}{!}{
    \begin{tabular}{lc|cc}
        \toprule[1pt]
        Model & TD & Pascal Context & Cityscape \\ \hline
        Simbaseline & & 40.1 & 30.0   \\ 
        Simbaseline & \checkmark & 42.6  & 34.7    \\ 
        \bottomrule[1pt]
        \vspace{-15pt}
    \end{tabular}
    }
\end{table}

\subsection{Qualitative Results}
\begin{figure*}[t]
    \centering
    \includegraphics[width=\linewidth]{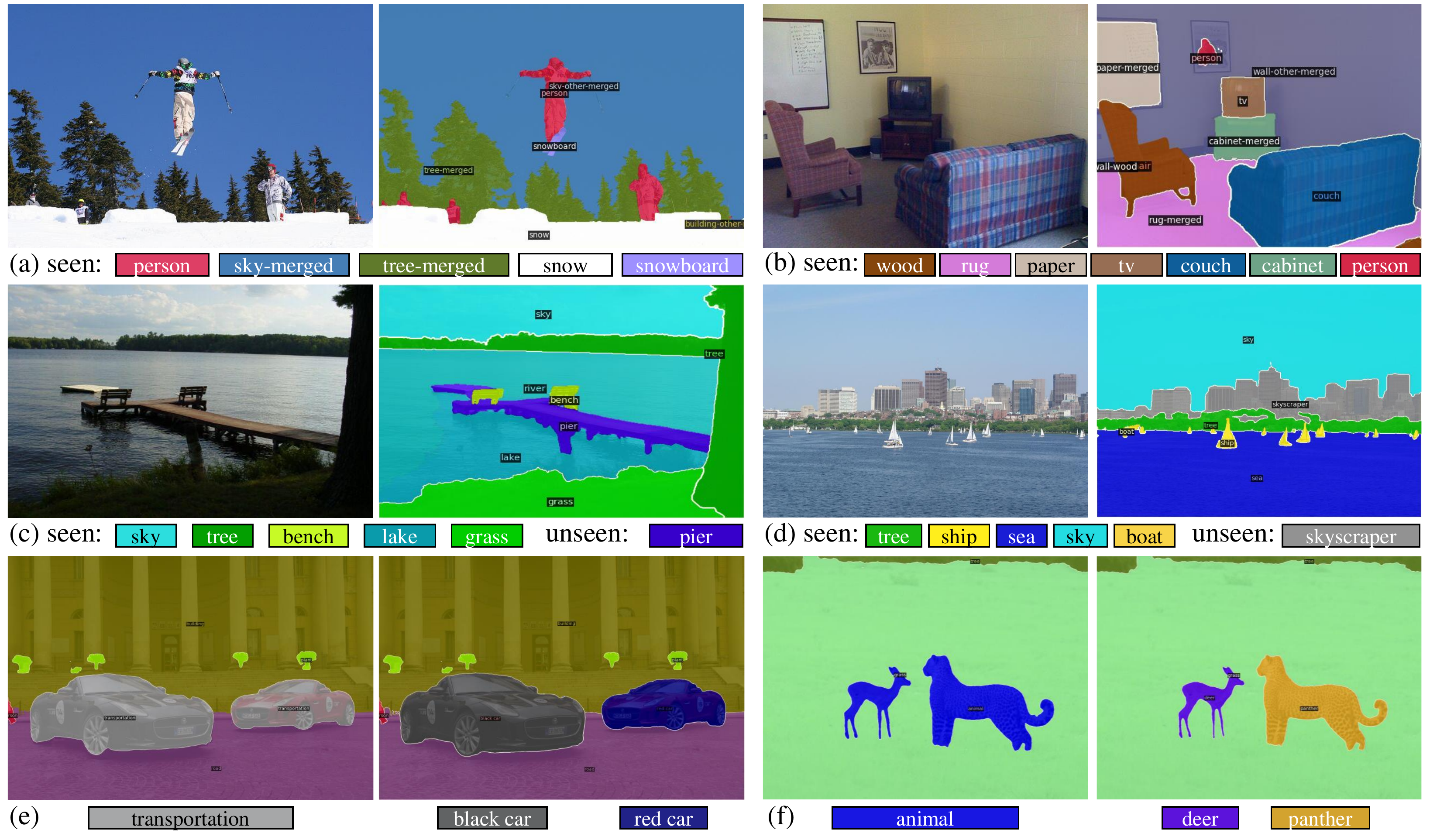}
    \caption{
        \textbf{Qualitative results.} (a) and (b) are evaluation results of COCO panoptic~\cite{coco} dataset, (c) and (d) are evaluation results of ADE20k-150~\cite{ade20k} dataset, (e) and (f) are inference results of different designed text prompts. For (a)-(d), categories of prediction are shown below, for (e) and (f), difference between twice text prompt inputs are shown below.
    }
    \label{fig:qualitative}
\end{figure*}
\cref{fig:qualitative} shows some visualization results of our method. 
From (c) and (d), we can see that our method is able to distinguish the regions of novel categories, \eg, ``pier" and ``skyscraper", from base categories. 
As one object can be described differently by multiple descriptions, for the same image, we also tested with different prompts to verify the open-vocabulary segmentation ability of our method. 
As shown in (e) and (f), our method can distinguish concepts of different granularities (\eg, ``transportation'' \vs ``black car'' or ``red car'', ``animal'' \vs ``deer'' or ``panther''). Note that none of these categories are used in COCO Panoptic~\cite{coco} and our text diversification strategy.

\section{Conclusion}
In this paper, we propose Global Knowledge Calibration that preserves the generalization ability during the training stage while enabling fast open-vocabulary image segmentation by abandoning the additional frozen CLIP during the inference stage. To broaden the text diversity, we leverage WordNet~\cite{wordnet} to avoid collapsing into particular known category names. We also propose text-guided knowledge distillation to utilize the well-aligned multi-modal space of CLIP~\cite{clip}. Extensive experiments on popular segmentation datasets demonstrate that our method outperforms previous methods in terms of performance and inference cost. To the best of our knowledge, we are the first to explore video open-vocabulary segmentation.

\smallskip \noindent \textbf{Limitations and Future Work.} 
We notice that our video open-vocabulary segmentation model still suffers from overfitting if we naïvely increase the number training iterations, resulting in performance degradation on novel categories. We will study these in future work.
% Despite being significantly faster than existing two-stage methods, the running speed is still far from real-time. We will study these in future work.

%%%%%%%%% REFERENCES
{\small
\bibliographystyle{ieee_fullname}
\bibliography{egbib}
}

\end{document}